\documentclass[letterpaper, 10 pt, conference]{ieeeconf}  

\IEEEoverridecommandlockouts                              

\usepackage{graphicx}
\usepackage{amsmath}
\usepackage{amssymb}
\usepackage{algorithm}
\usepackage{algpseudocode}
\usepackage{algorithmicx}
\usepackage{booktabs}
\usepackage{multirow}
\usepackage{makecell}
\usepackage{url}
\usepackage{cite}
\usepackage{xcolor}
\usepackage{colortbl}

\let\OLDthebibliography\thebibliography
\renewcommand\thebibliography[1]{%
  \OLDthebibliography{#1}%
  \setlength{\itemsep}{0pt plus 0.2ex}%
  \setlength{\parsep}{0pt}%
}

\title{SenseExpo: \textbf{S}patial \textbf{E}xploration and \textbf{N}avigation via \textbf{S}cene \textbf{E}stimation from \textbf{Ex}peditious \textbf{P}redictive \textbf{O}perators}

\author{Haojia Gao$^{1,\dagger}$, Haohua Que$^{2,5,\dagger}$, Mingkai Liu$^{3}$, Jiayue Xie$^{1}$, Hoiian Au$^{1}$, Yusen Qin$^{4}$,
Qian Zhang$^{1,5}$, \\
Jiajun Sun$^{1}$, Weihao Shan$^{1}$, Tianle Zhu$^{2}$,
Handong Yao$^{2,*}$, Fei Qiao$^{1,*}$%
\thanks{$^{1}$Tsinghua University, $^{2}$University of Georgia, $^{3}$Peking University, $^{4}$D-Robotics, $^{5}$Infinity Robotics. $^{\dagger}$Equal contribution (co-first authors). $^{*}$Corresponding authors: Handong Yao (\texttt{handong.yao@uga.edu}; first corresponding author), Fei Qiao (\texttt{qiaofei@tsinghua.edu.cn}; second corresponding author).}%
}

\begin{document}

\maketitle
\thispagestyle{empty}
\pagestyle{empty}

\begin{abstract}
	We present \textbf{SenseExpo}, a lightweight single-robot exploration framework that integrates a compact map prediction network into a frontier-based strategy. SenseExpo addresses two long-standing challenges in classical methods---high computational overhead and poor environmental generalization. Our prediction network combines Generative Adversarial Networks (GANs), Transformers, and Fast Fourier Convolution (FFC) to achieve a remarkably small footprint of only 709K parameters. Despite its compactness, SenseExpo outperforms U-Net (24.5M) and LaMa (51M) on the KTH dataset, achieving PSNR 9.026 and SSIM 0.718, representing a 38.7\% PSNR gain over LaMa. Cross-domain evaluation further verifies strong generalization with an FID of 161.55 on HouseExpo. In exploration experiments, SenseExpo reaches target coverage 67.9\% faster on KTH and 77.1\% faster on MRPB~1.0 than a MapEx-style global obstacle-prediction baseline under the same simulator; because the methods predict different map semantics, this comparison evaluates planning utility rather than a direct predictor ranking. Implemented as a plug-and-play ROS (Robot Operating System) node, our framework integrates with existing navigation stacks, providing an efficient solution for resource-constrained robotic systems.
\end{abstract}

\section{INTRODUCTION}
\label{sec:intro}

Autonomous exploration systems play a pivotal role in numerous fields, such as planetary exploration~\cite{vogele2024robotics} and environmental monitoring~\cite{dunbabin2012robots}, enabling robots to operate in unknown or partially known environments without human intervention.
However, one of the significant challenges in these settings is the ability to efficiently explore and map the surroundings in real-time.

\begin{figure*}[thpb]
	\centering
	\includegraphics[width=\textwidth]{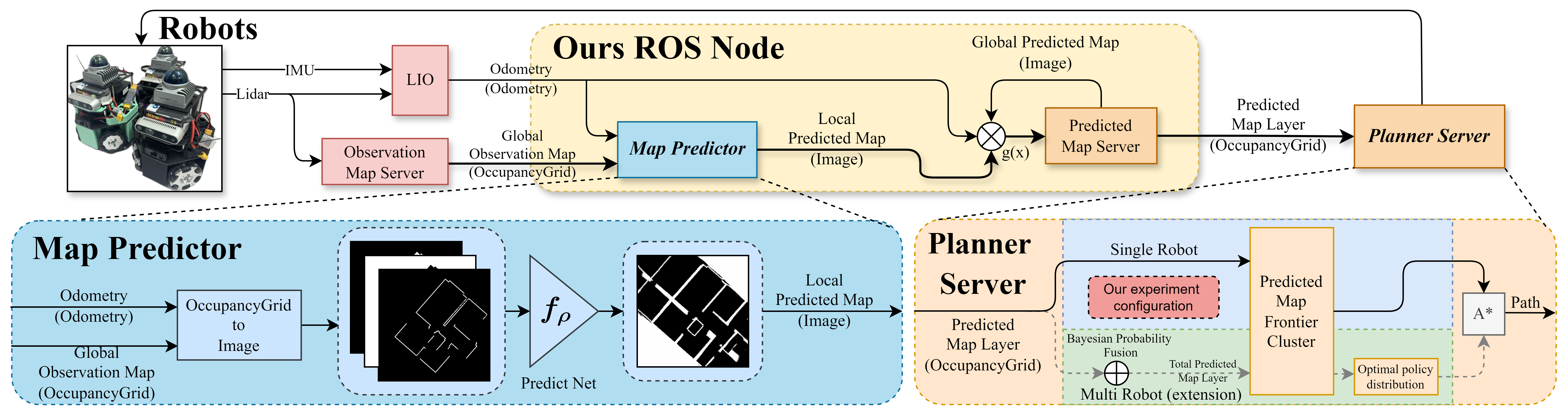}
	\caption{ \textbf{The Overview of \emph{SenseExpo}.} The Map Predictor uses odometry and global observation map obtained from the robot to produce local predicted map. The Predicted Map Server then concatenates the local predicted map and the global predicted map to form a predicted map layer. For a potential multi-robot extension, the Planner Server would be responsible for fusing maps from each robot (if there are multi robots) into a total predicted map layer and outputs the path to the navigation point.}
	\label{pipeline}
\end{figure*}

For robots, the exploration of unknown environments remains a complex challenge due to the limited sensor field of view (FOV) and the computational cost of generating maps in real-time. \textit{So how to make robots explore smarter with less cost?} Biological systems hold the key. Unlike robots, humans and animals navigate unknown spaces by building mental maps that extend far beyond their immediate view, allowing them to make better decisions and anticipate future states~\cite{tolman1948cognitive,epstein2017cognitive,madl2015computational}.
So we think robots also can obtain such `memory' ability by pre-trained models to explore more efficiently.
There is already some related work in this field. Shrestha et al.~\cite{shrestha2019learned} learned occupancy-map completion to extend frontier-based exploration, MapEx~\cite{ho2024mapex} used LaMa~\cite{suvorov2022resolution} to predict global map structures, and Katyal et al.~\cite{katyal2019uncertainty} used a U-Net~\cite{ronneberger2015u} network to predict maps under uncertainty. Although progress has been made, current prediction-based exploration methods still face high computational overhead, limited cross-domain generalization, and integration costs on resource-constrained robots.

To address these challenges, we introduce SenseExpo. While building upon the well-established frontier-based exploration paradigm~\cite{yamauchi1997frontier}, the core novelty of \emph{SenseExpo} lies in its map prediction model\textemdash a lightweight (only 709k parameters at minimum) yet powerful network that overcomes the efficiency and deployment challenges limiting previous prediction-based methods.
Unlike previous methods that rely on global map inputs and local feature aggregation, our model operates directly on partial observations captured by the robot's onboard sensors, processing only local occupancy maps centered around its current position.
\emph{SenseExpo} proposes an efficient predictive network integrating GANs~\cite{goodfellow2014generative}, Transformer~\cite{vaswani2017attention}, and Fast Fourier Convolution (FFC)~\cite{chi2020fast}, overcoming the limitations of traditional local feature aggregation.
The Transformer models long-range dependencies through multi-head attention, while adversarial training employs a lightweight discriminator to encourage plausible spatial layouts in uncertain regions. FFC employs a dual-branch structure: the local branch uses convolutional operations to extract detailed features, while the global branch captures contextual information through frequency-domain transformations. This architecture reduces computational cost while enhancing global structural reasoning.

Our main contributions are threefold. First, we introduce \emph{SenseExpo}, an approach that integrates a novel lightweight prediction network into an autonomous exploration system, enabling deployment on resource-constrained devices. Second, we propose a local map prediction network that significantly improves robustness and accuracy on unseen datasets. Third, we package \emph{SenseExpo} as a plug-and-play ROS node and demonstrate efficient exploration against frontier and MapEx-style baselines across multiple datasets~\cite{aydemir2012can,wen2021mrpb}.

\section{RELATED WORK} \label{sec:related_works}

\begin{figure*}[thpb]
	\centering
	\includegraphics[width=\textwidth]{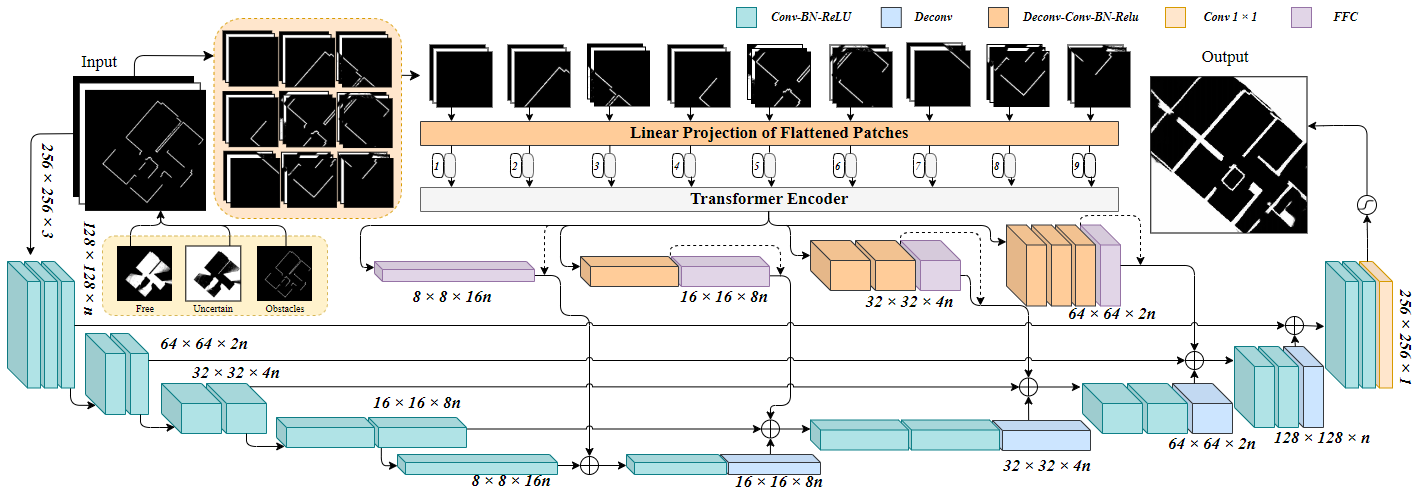}
	\caption{\textbf{Complete Architecture of the Map Prediction Network.} The input, which has 3 channels (free, uncertain, obstacles), is processed through the network to produce a single-channel grayscale image output, where each pixel value represents the probability of occupancy. Simultaneously, input patches are passed through a Transformer Encoder and then Fast Fourier Convolution, with the outputs concatenated to the feature maps in the U-Net.}
	\label{fig3}
\end{figure*}

\subsection{Autonomous Exploration} \label{sec:related_works:autonomous_exploration}

Conventional strategies typically identify frontiers~\cite{yamauchi1997frontier}, which is the boundaries separating explored areas from uncharted regions. This idea uses the closest frontier as next best goal to systematically guide incremental map expansion~\cite{yamauchi1997frontier, holz2010evaluating,  dornhege2013frontier}.
Frontier-driven strategies are particularly effective in large-scale environments, detecting unmapped regions within the global map.
Another category of popular methodologies include random sampling-based exploration strategies, which utilize random sampling to identify unexplored areas. Recent studies include a few methods based on the Rapidly-exploring Random Tree (RRT) algorithm~\cite{lavalle1998rapidly} and Rapidly-exploring Random Graph (RRG)~\cite{karaman2011sampling}.
These methods build probabilistic roadmaps through systematic sampling and choose the maximum-information edge as the exploration trajectory. For instance, Hollinger et al.~\cite{hollinger2014sampling} proposed sampling-based motion planners that maximize an information-quality metric, and Steinbrink et al.~\cite{steinbrink2021rapidly} used an RRG with Next-Best-View (NBV) evaluation.
Moreover, hybrid methods combine both frontier-based and sampling-based planning. This mix uses the strength of each approach, as demonstrated in recent studies~\cite{selin2019efficient, respall2021fast}.
Other methods include topological exploration~\cite{choset2000sensor, acar2002sensor} and information-theoretic approaches~\cite{bourgault2002information, stachniss2005information}, which use multi-step planning to maximize information gain.

\subsection{Map Prediction}
Anticipating unknowns in uncharted territory is a critical capability for autonomous robots. The foundational work by Thrun et al.~\cite{thrun1998probabilistic} established a probabilistic framework for generating the most likely maps from data. The KTH~\cite{aydemir2012can} and HouseExpo~\cite{li2020houseexpo} datasets are two milestone indoor benchmarks providing standardized evaluation for learning-based map prediction.
Recently, to leverage structural predictability in unknown environment, there has been some work that uses deep learning to predict unknown regions of maps. For instance, UPEN~\cite{georgakis2022uncertainty} presented an uncertainty-driven method to generate occupancy maps and
IG-Hector~\cite{shrestha2019learned} used a predicted map to denote how much area can be observed from a viewpoint. Shrestha et al.~\cite{shrestha2019learned} are the most closely related to our planning idea because they also use learned map prediction to extend frontier-based exploration. SenseExpo differs by emphasizing local sensor-centered prediction, a compact hybrid network, and ROS deployment; therefore, our comparison to U-Net-style models should be read as an architectural baseline rather than a reproduced head-to-head implementation of that system. Particularly, MapEx~\cite{ho2024mapex} generated multiple predicted maps from the observed information, demonstrating superior exploration performance and topological understanding compared to the first two.
Furthermore, with the advancement of multi-agent systems, map prediction methods have found extensive applications in this domain, and recent extensions leverage diffusion models (DMs) for probabilistic spatial forecasting~\cite{smith2018distributed}.

\subsection{GAN for Robotic} \label{sec:related_works:gan_for_robotic}
Since their introduction by Ian Goodfellow in 2014~\cite{goodfellow2014generative}, GANs have been widely applied in image-to-image translation and image generation~\cite{zhou2023gan}, and have gained traction in robotics. For instance, Ren et al.~\cite{ren2020learning} learned inverse kinematics and dynamics of a robotic manipulator, and Zhang et al.~\cite{zhang2021generative} used GANs to improve the initial-solution quality and convergence speed in path planning.
While prediction models generate global maps well, their global reasoning paradigm suffers from two critical limitations: (1) computational complexity exceeding the resource constraints of edge devices, and (2) distortion-prone predictions on novel scene layouts. Existing implementations also rely on proprietary middleware, complicating ROS integration and real-world deployment. Our contributions develop new methodologies to address these limitations.
In our setting, the GAN objective is used as an auxiliary structural prior rather than a replacement for supervised learning: the available ground truth is still enforced through $L_1$ and ResNetPL losses, while the discriminator penalizes locally plausible but globally inconsistent completions in uncertain regions.

\section{PROBLEM DEFINITION} \label{sec:problem}

\paragraph{Notation.}
Throughout this paper, we use the following symbols consistently:
$M$ denotes the ground-truth global occupancy map;
$M_{\text{obs}}^{t}$ represents the observation-only map at time $t$;
$M_{\text{pred}}^{t}$ denotes the predicted probability map produced by our model;
and $m_{\text{obs}}^{t}, m_{\text{pred}}^{t}$ are the corresponding local patches.
All probabilities $p(\text{occupied})$ lie within $[0,1]$.

Consider an unknown environment modeled as a continuous 2D space $E \subseteq \mathbb{R}^{2}$. The ground truth of this environment is represented by an occupancy grid map, an unknown 2D discrete matrix $M \in \{0, 1\}^{H \times W}$.
This matrix is a discretization of the continuous environment $E$, where the space is divided into a grid of uniform (square) cells. Each cell in $M$ corresponds to a specific area in $E$. A value of `1' in a cell signifies that the corresponding area is \textbf{occupied} by an obstacle, while a value of `0' signifies it is \textbf{free} space.
In this environment, $n$ mobile robots $\{R_{i}\}_{i=1}^{n}$ are deployed, and each robot satisfies the following conditions:
\begin{itemize}
	\item The pose of each robot $R_{i}$ at time $t$ is denoted as $P_{i}^{t} \in E$, which exists in the continuous space $E$. At each time step $\Delta t$, it moves a fixed distance $L$ along the planned path, such that the movement satisfies the constraint:
	      \begin{equation}
		      \|P_{i}^{t+\Delta t} - P_{i}^{t}\|_{2} = L
	      \end{equation}

	\item Each robot $R_{i}$ is equipped with a laser scanner with an effective radius $r$, and at time $t$, it can obtain a local observation of the obstacle map $m_{i}^{t} = S(E, P_{i}^{t}, r, \delta)$. These observations are used to build a \textbf{global observation map $M_{i}^{t}$}, which contains only ground-truth information from the sensor. The side length of the square local observation map is $\sigma = \delta \cdot 2r$, where $\delta$ is a unitless scaling factor that determines the size of the observation window relative to the sensor's diameter.

	\item Each robot $R_{i}$ is also equipped with a map prediction network $f_{\rho}: m_{i}^{t} \rightarrow \hat{m}_{i}^{t}$ where $\hat{m}_{i}^{t}$ is the predicted local obstacle map of robot $R_{i}$ at time $t$, and the network has a parameter size of $\rho$.

	\item Each robot maintains a \textbf{global predicted map} $\hat{M}_{i}^{t}$, which is constructed by concatenating the local predicted maps $\{\hat{m}_{i}^{\tau}\}_{\tau=0}^{t}$ over time.
	      This map layer consists solely of predictions generated by the model.
\end{itemize}

The robot moves in continuous space, while its perception and planning representation use the discrete grid map $M$. Although the notation admits multiple robots, the evaluation and claims in this paper focus on the foundational single-robot ($n=1$) case; multi-robot fusion is left as a future extension.

The optimization goal is to minimize the model parameter size $\rho$ and the total exploration time $T$ while ensuring that the total predicted map $\hat{M}_{\text{Total}}^{T}$ is as close as possible to the ground truth map $M$:
\begin{equation}
	\min(\theta_{1} \cdot \rho + \theta_{2} \cdot T + \theta_{3} \cdot \|M - \hat{M}_{\text{Total}}^{T}\|_{1})
\end{equation}
Where:
\begin{itemize}
	\item When there is only one robot ($n=1$), the total predicted map is $\hat{M}_{\text{Total}}^{T} = \hat{M}_{0}^{t}$.
	\item For a potential multi-robot extension ($n \neq 1$), the total predicted map can be written as $\hat{M}_{\text{Total}}^{T} = \mathcal{F}(\{\hat{M}_{i}^{T}\}_{i=1}^{n})$, where $\mathcal{F}$ is a map fusion operator. We do not evaluate this extension in the present work.
\end{itemize}

\section{METHODOLOGY} \label{sec:methods}

We present the overall pipeline for efficient exploration in unknown environments (Fig.~\ref{pipeline}). First, we introduce our lightweight local map prediction model and its GAN-based training procedure. Next, we detail the frontier-based exploration strategy, focusing on the single-robot case.

\subsection{Map Prediction Model} \label{sec:methods:map_prediction_model}
\begin{figure}[thpb]
	\centering
	\includegraphics[width=0.46\textwidth]{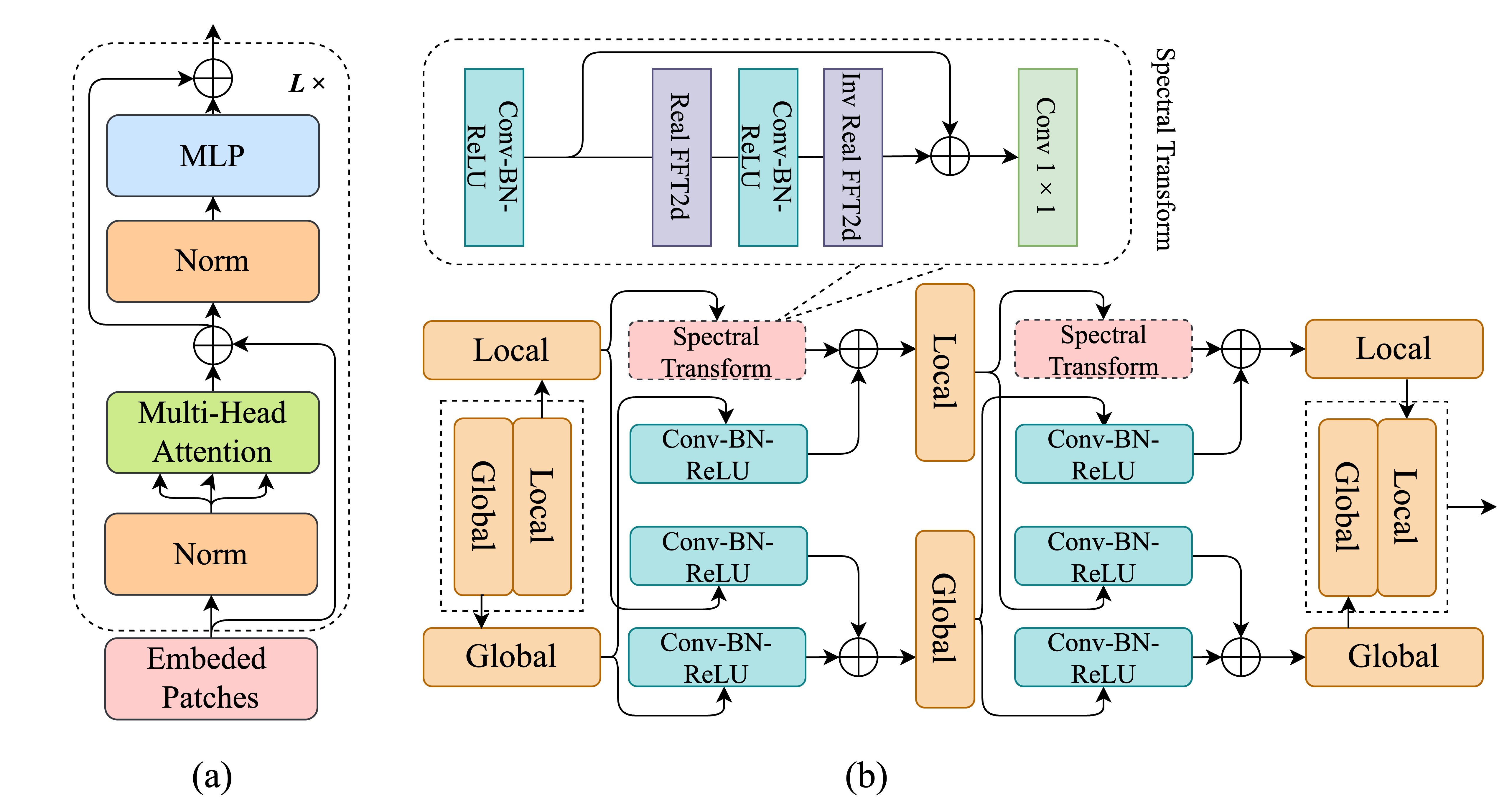}
	\caption{\textbf{Model Architecture.} (a) The Transformer Encoder extracts high-level features from embedded patches. (b) The spectral transform module combines Conv-BN-ReLU and FFT2d layers for local and global features.}
	\label{fig4}
\end{figure}

Currently, most map prediction algorithms face deployment challenges due to their substantial parameter sizes and poor generalization, especially on robots with limited training data and in highly variable environments~\cite{zwecher2020deep}. Most map prediction algorithms are based on U-Net~\cite{ronneberger2015u, shrestha2019learned, zwecher2020deep, katyal2019uncertainty}, or image reconstruction networks like LaMa~\cite{suvorov2022resolution, ho2024mapex, spinos2025map}. Our approach uses a local map prediction method that improves the model's generalization capability, allowing it to maintain a certain level of accuracy even in unknown environments while significantly reducing the number of parameters, effectively overcoming the aforementioned limitations. Our model architecture is shown in Fig.~\ref{fig3} and Fig.~\ref{fig4}.
We reduce the number of channels in each layer of the U-Net~\cite{ronneberger2015u} network and remove redundant convolutional layers, while keeping the Deconv-Conv-BN-ReLU blocks aligned with the encoder-decoder resolution hierarchy so that each upsampling stage restores one spatial scale. The Transformer Encoder~\cite{vaswani2017attention, dosovitskiy2020image} is selected because attention can directly connect distant corridor and room structures without stacking many dilated convolutions. FFCs~\cite{chi2020fast} are selected because their spectral branch captures low-frequency global layout cues more parameter-efficiently than multi-branch dilation modules such as ASPP. Additionally, we add Dropout Layers with a rate of 0.5 after multiple convolution operations to enhance robustness and generalization. These improvements reduce the parameter count while improving map prediction accuracy.

To represent the three distinct states of the local map, we encode it into three binary channels. The \textbf{`Free'} and \textbf{`Obstacle'} channels mark cells that are explicitly known from sensor data. As can be inferred from Fig.~\ref{fig5}, the \textbf{`Uncertain'} channel is then determined as the logical complement, marking all cells that are neither known to be free nor occupied ($\textit{Uncertain} = \neg(\textit{Free} \lor \textit{Obstacle})$).

\subsection{Model Training} \label{sec:methods:model_training}

\begin{figure}[thpb]
	\centering
	\includegraphics[scale=0.43]{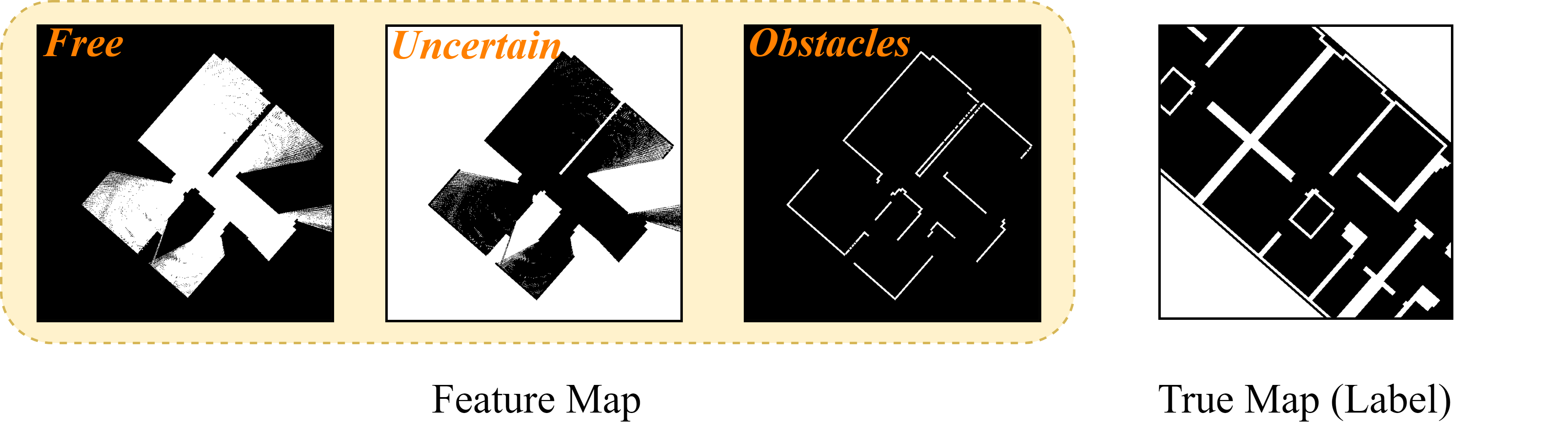}
	\caption{\textbf{The example of dataset.} The True Map is encoded into three channels, free, uncertain and obstacles.}
	\label{fig5}
\end{figure}

We modified the PseudoSLAM~\cite{li2020houseexpo} designed by Li et al. to construct a virtual environment, and followed the dataset collection procedure proposed in KTH~\cite{aydemir2012can}. The dataset collects features as local observation maps, which are encoded into three channels, with labels representing the ground truth of the local maps as shown in Fig.~\ref{fig5}, used for model training.

During training, we use a conditional PatchGAN-style encoder-decoder discriminator. Its input is the concatenation of the three-channel observed local map and either the one-channel ground-truth local map (real sample) or the generated prediction (fake sample). The discriminator contains four downsampling blocks with 32, 64, 128, and 256 channels, mirrored by three upsampling blocks and a final sigmoid patch classifier. We use the vanilla binary cross-entropy GAN objective, masked to the unknown part of the local map so that adversarial feedback focuses on hallucinated regions rather than already observed cells. Furthermore, we introduce the ResNet50-based perceptual loss network (ResNetPL)~\cite{suvorov2022resolution} during training, and the specific calculation is as follows:

\begin{equation}
	L_{\text{ResNetPL}} = \sum_i \| \psi_i(x) - \psi_i(\hat{x}) \|^2
\end{equation}

where \( \psi_i(*) \) is the feature map of the \( i \)-th layer in the ResNet50 with dilated convolutions~\cite{zhou2019semantic}, which helps the model focus more on the local region.

Additionally, we incorporate an $L_1$ loss to make the generated map maintain clearer edges and details. Hence, our total loss function is calculated as:

\begin{equation}
	L_{\text{Total}} = \omega_1 L_{\text{Adv}} + \omega_2 L_{\text{ResNetPL}} + \omega_3 L_1
\end{equation}
The adversarial term is therefore used to improve the structural realism of predicted free/occupied regions, while $L_1$ and ResNetPL keep the prediction anchored to the supervised map labels.

\subsection{Autonomous Exploration based on Map Prediction} \label{sec:methods:autonomous_exploration}
Our exploration strategy leverages the classical frontier-based~\cite{yamauchi1997frontier} search algorithm  to ensure systematic coverage. The primary innovation, however, is not in redefining this paradigm, but in how our novel map predictor (Fig.~\ref{fig3}) provides high-quality, long-range predictive information.
Our predictor produces a single-channel probability map $M_{pred}^t$, where each pixel is the probability of occupancy $p(\text{occupied})\in[0,1]$ (consistent with Fig.~\ref{fig3}).
We convert $M_{pred}^t$ into a planning map via thresholds:
\begin{equation}
	\mathcal{C}(p)=
	\begin{cases}
		\text{free},      & p < 0.10,            \\
		\text{uncertain}, & 0.10 \le p \le 0.90, \\
		\text{obstacle},  & p > 0.90.
	\end{cases}
\end{equation}
The thresholds are intentionally conservative: only high-confidence free cells are used to expand the planning frontier, while the wide uncertain band absorbs noisy probability fluctuations near walls. In practice, small perturbations of these thresholds mainly change the width of the candidate frontier band; collision safety still depends on the observation map used by the navigation stack. A systematic threshold-sensitivity study is beyond the present camera-ready revision and is left for future work.

Let $M_{\text{obs}}^{t}$ be the observation-only map and $M_{\text{pred}}^{t}$ the predicted map after binarization above.
We define frontiers as boundary points between \emph{observed free} cells in $M_{\text{obs}}^{t}$ and \emph{predicted uncertain} cells in $M_{\text{pred}}^{t}$.

\begin{figure}[thpb]
	\centering
	\includegraphics[scale=0.35]{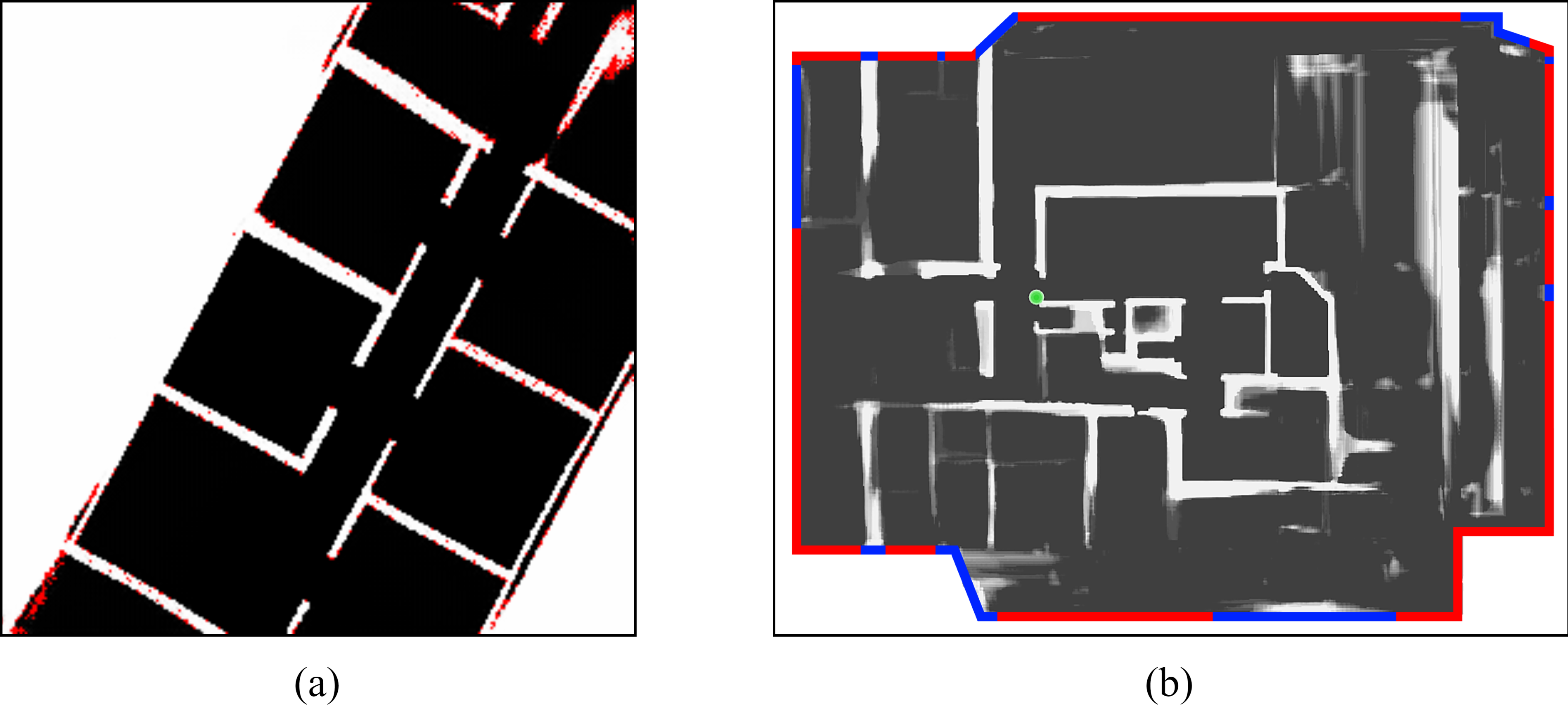}
	\caption{\textbf{Comparison of traditional and proposed methods.} (a) Uncertain regions concentrated at obstacle edges in the traditional approach; (b) Boundary points extracted from the free region in the proposed method.}
	\label{fig6}
\end{figure}

This transforms goal selection, substantially improving exploration efficiency over using only currently observed data.
The traditional Frontier-based algorithm searches for boundary points between the uncertain and free areas. If we define thresholds to partition the free, uncertain, and obstacle regions, the uncertain area will mostly be concentrated at the edges of the obstacles, as shown in Fig.~\ref{fig6}(a). This makes it unsuitable for use as a navigation point selection area.

To address the problem mentioned above, we designed a frontier-based search algorithm~\cite{yamauchi1997frontier} based on local predicted maps.
To accommodate potential multi-robot scenarios, after each robot updates its local predicted map \( \hat{m}_i^t \), a fusion mechanism such as Bayesian Updating could be used to obtain the total predicted map \( \hat{M}_{\text{Total}}^t \). For the single-robot case, which is the focus of our experiments, the total predicted map is simply the local predicted map \( \hat{m}_i^t \). The boundary points between the free region (red region) in \( \hat{M}_{\text{Total}}^t \) are extracted as shown in Fig.~\ref{fig6}(b).

We associate each frontier with its corresponding area, allowing the robot to focus on exploring larger regions in the unexplored areas.
The exploration process continues iteratively, with the robot moving to the highest-utility frontier at each step.
Exploration terminates only when no new frontiers exist on $M_{\text{obs}}^{t}$ (the observed map), not on $M_{\text{pred}}^{t}$.
This termination condition is explicitly handled in our goal selection algorithm (Alg.~\ref{alg:goal_selection}).

\begin{algorithm}[t]
	\caption{Frontier-based Goal Selection (Single Robot)}
	\label{alg:goal_selection}
	\small
	\begin{algorithmic}[1]
		\Require Observed map $M_{\text{obs}}^{t}$, Predicted map $M_{\text{pred}}^{t}$, Current pose $P^{t}$
		\Ensure Next goal $G^{t}$, or \textit{ExplorationComplete}
		\State $F_{\text{pts}} \gets \text{BoundaryBetween}(\text{ObsFree}(M_{\text{obs}}^{t}), \text{PredUnc}(M_{\text{pred}}^{t}))$
		\If{$F_{\text{pts}}$ is empty}
		\State \Return $(\text{null},~\text{ExplorationComplete})$
		\EndIf
		\State $F_{\text{clusters}} \gets \text{DBSCAN}(F_{\text{pts}},~\epsilon,~min\_pts)$
		\For{each cluster $C_j$ in $F_{\text{clusters}}$}
		\State $\mu_j \gets \text{Centroid}(C_j)$, $size_j \gets \text{Area}(C_j)$
		\State $cost_j \gets \| \mu_j - P^{t} \|_{2}$, $U_j \gets size_j - w \cdot cost_j$
		\EndFor
		\State $j^\star \gets \arg\max_{j} U_j$, $G^{t} \gets \mu_{j^\star}$
		\State \Return $(G^{t},~\neg \text{ExplorationComplete})$
	\end{algorithmic}
	\vspace{-1mm}
\end{algorithm}

\section{EXPERIMENTS AND DISCUSSION} \label{sec:experiment}

\subsection{The Evaluation of Map Prediction} \label{sec:experiment:the_evaluation_of_map_prediction}

\subsubsection{Experiment Settings and Metrics} \label{sec:experiment:the_evaluation_of_map_prediction:a}
Our training environment included an AMD EPYC 9654 CPU, 1 TB RAM, eight NVIDIA RTX 4090 GPUs (24GB VRAM each), Python 3.12, PyTorch 2.4, and CUDA 12.4.
We use the KTH~\cite{aydemir2012can} dataset (30,000 local-map samples); each sample has a \(256 \times 256 \times 3\) feature array (local observation map, Sec.~\ref{sec:methods:map_prediction_model}) and a \(256 \times 256 \times 1\) label (local ground truth), split into 24,000 training and 6,000 testing samples. We compare U-Net~\cite{ronneberger2015u}, LaMa~\cite{suvorov2022resolution}, and our model at different scales.
The evaluation employs: parameter count (network lightweightness); Peak Signal-to-Noise Ratio (PSNR), for prediction quality; Structural Similarity (SSIM)~\cite{wang2004image}, for structural consistency between $\hat{m}$ and $m$; Learned Perceptual Image Patch Similarity (LPIPS)~\cite{zhang2018unreasonable}, for perceptual similarity via pre-trained VGG~\cite{simonyan2014very} and AlexNet~\cite{krizhevsky2012imagenet} on ImageNet~\cite{deng2009imagenet}; and Fr\'{e}chet Inception Distance (FID)~\cite{bynagari2019gans,Seitzer2020FID}, for feature-distribution distance between $\hat{m}$ and $m$.
Although these metrics originate from image reconstruction, they are useful for occupancy maps because the prediction target is a dense grid: PSNR reflects cell-wise errors, SSIM reflects local structural consistency, and LPIPS/FID serve only as auxiliary distributional indicators. We therefore report navigation coverage and exploration time separately rather than treating FID or LPIPS as direct navigation metrics.
The corresponding calculation formulas are as follows:

\begin{equation}
	\mathrm{PSNR}(\hat{m}, m)=10 \log_{10}\!\left(\frac{1}{\mathrm{MSE}(\hat{m}, m)}\right).
\end{equation}
where \(\text{MSE}(\hat{m}, m)\) is the Mean Squared Error between the predicted and ground truth maps.

\begin{equation}
	\text{SSIM}(\hat{m}, m) = \frac{(2v_{\hat{m}} v_m + C_1)(2\sigma_{\hat{m}m} + C_2)}{( v_{\hat{m}}^2 + v_m^2 + C_1)(\sigma_{\hat{m}}^2 + \sigma_m^2 + C_2)}
\end{equation}
where \( v_{\hat{m}}, v_m \) are the mean intensities, \( \sigma_{\hat{m}}^2, \sigma_m^2 \) the variances, \( \sigma_{\hat{m}m} \) the covariance (capturing structural similarity), and \( C_1, C_2 \) stabilizing constants.

\begin{equation}
	\mathrm{LPIPS}(\hat{m}, m)=\sum_{l}\left\| \varphi_{l}(m)-\varphi_{l}(\hat{m}) \right\|_2.
\end{equation}
where \(\varphi_l(*)\) represents the feature maps extracted from deep networks.

\begin{equation}
	\text{FID}(\hat{m}, m) = \| \mu_{\hat{m}} - \mu_m \|_2^2 + \text{Tr}(\Sigma_{\hat{m}} + \Sigma_m - 2(\Sigma_{\hat{m}} \Sigma_m)^{1/2})
\end{equation}
where \(\mu_{\hat{m}}, \mu_m\) and \(\Sigma_{\hat{m}}, \Sigma_m\) represent the means and covariances of the \( \hat{m} \) and the \( m \) features, respectively.

\subsubsection{Experimental Results}
\begin{figure}[thpb]
	\centering
	\includegraphics[width=0.35\textwidth]{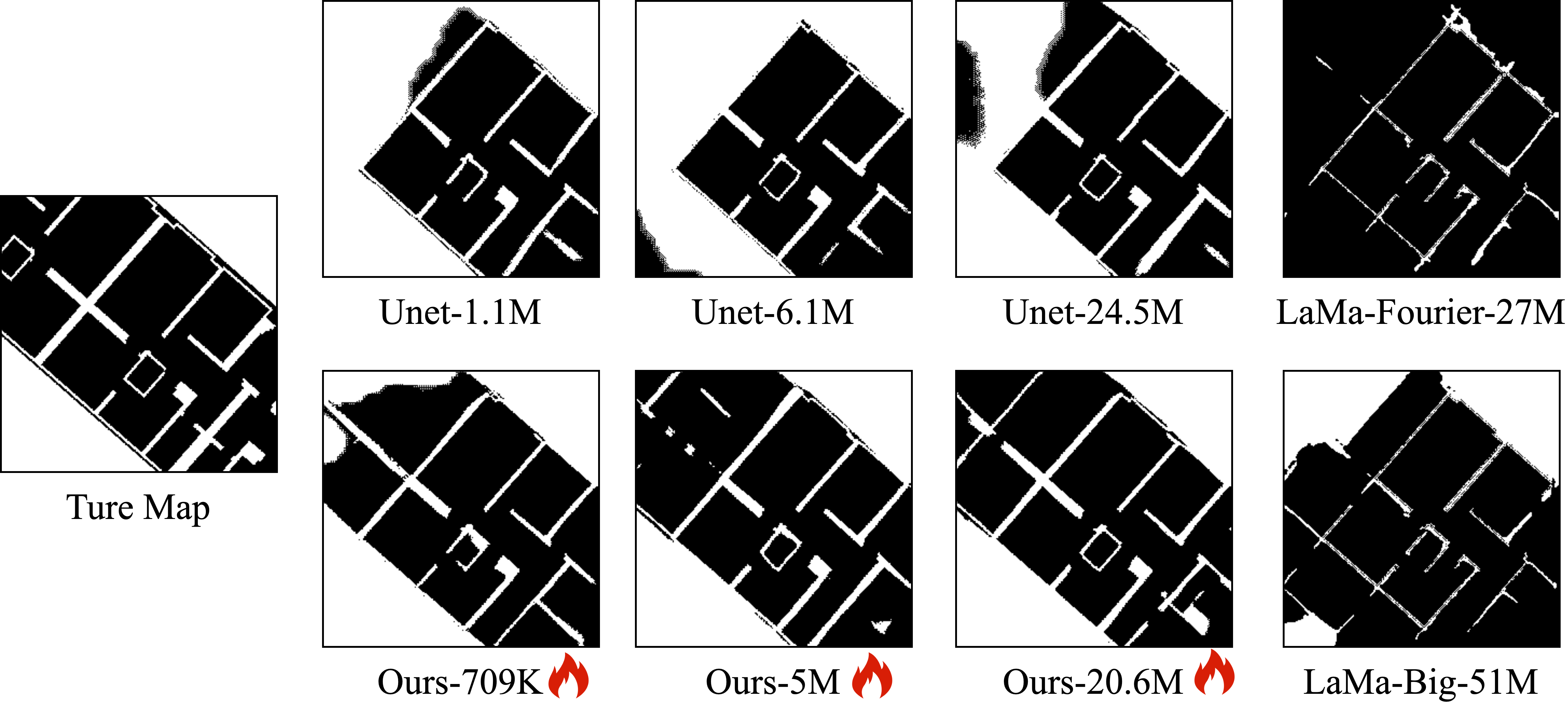}
	\caption{\textbf{The example of map prediction.} Comparison of ground truth map and predicted maps. Our models outperform U-Net, which tends to cover a more limited area, as well as the LaMa model, whose predictions show structural inaccuracies and lower clarity.}
	\label{fig7}
	\vspace{-10pt}
\end{figure}

The comparison results are shown in Fig.~\ref{fig7} and Tab.~\ref{tab:comprehensive_comparison}, evaluating prediction quality, structural and perceptual similarity, and feature-distribution distance. Although larger models generally perform better, our lightweight model (709K parameters) already reaches a PSNR of 9.026 and an SSIM of 0.718, surpassing both the small and large U-Net variants despite being over 30$\times$ smaller. Increasing capacity to 20.6M parameters further improves PSNR/SSIM to 9.209/0.731 and reduces FID from 42.35 to 33.16, confirming the effectiveness and scalability of our design across parameter regimes.

\begin{table}[t]
	\centering
	\scriptsize
	\setlength{\abovecaptionskip}{2pt}
	\setlength{\belowcaptionskip}{0pt}
	\setlength{\tabcolsep}{3pt}
	\caption{Comparison of Model Performance and Robustness}
	\label{tab:comprehensive_comparison}
	\renewcommand{\arraystretch}{0.85}
	\begin{tabular*}{\linewidth}{@{\extracolsep{\fill}}ccccccc@{}}
		\toprule
		\textbf{Exp.} & \textbf{Size}           & \textbf{Network} & \textbf{Params}$\downarrow$        & \textbf{PSNR}$\uparrow$            & \textbf{SSIM}$\uparrow$            & \textbf{FID}$\downarrow$           \\
		\midrule
		\multirow{8}{*}{\rotatebox{90}{Perf. Comp.}}
		& \multirow{2}{*}{S}      & U-Net            & 1.1M                               & 7.707                              & 0.670                              & 56.90                              \\
		&                         & \textbf{Ours}    & \cellcolor{green!20}\textbf{709K}  & \cellcolor{green!20}\textbf{9.026} & \cellcolor{green!20}\textbf{0.718} & \cellcolor{green!20}\textbf{42.35} \\
		\cmidrule(lr){2-7}
		& \multirow{3}{*}{M}      & LaMa             & 27.0M                              & 7.428                              & 0.654                              & 169.8                              \\
		&                         & U-Net            & 6.1M                               & 8.267                              & 0.695                              & 47.10                              \\
		&                         & \textbf{Ours}    & \cellcolor{green!20}\textbf{5M}    & \cellcolor{green!20}\textbf{9.189} & \cellcolor{green!20}\textbf{0.724} & \cellcolor{green!20}\textbf{38.05} \\
		\cmidrule(lr){2-7}
		& \multirow{3}{*}{L}      & Big LaMa         & 51.0M                              & 6.508                              & 0.608                              & 117.9                              \\
		&                         & U-Net            & 24.5M                              & 8.539                              & 0.711                              & 41.17                              \\
		&                         & \textbf{Ours}    & \cellcolor{green!20}\textbf{20.6M} & \cellcolor{green!20}\textbf{9.209} & \cellcolor{green!20}\textbf{0.731} & \cellcolor{green!20}\textbf{33.16} \\
		\midrule
		\multirow{5}{*}{\rotatebox{90}{Pred. Robust.}}
		& \multirow{2}{*}{Global} & LaMa             & 27.0M                              & 3.597                              & 0.428                              & 409.8                              \\
		&                         & Big LaMa         & 51.0M                              & 3.824                              & 0.444                              & 397.4                              \\
		\cmidrule(lr){2-7}
		& \multirow{3}{*}{Local}  & LaMa             & 27.0M                              & 4.296                              & 0.506                              & 261.4                              \\
		&                         & Big LaMa         & 51.0M                              & 5.171                              & 0.570                              & 188.1                              \\
		&                         & \textbf{Ours}    & \cellcolor{green!20}\textbf{709K}  & \cellcolor{green!20}\textbf{5.673} & \cellcolor{green!20}\textbf{0.605} & \cellcolor{green!20}\textbf{161.5} \\
		\bottomrule
	\end{tabular*}
\end{table}

Notably, our 709K model outperforms the Big LaMa-Fourier model (51.0M) with a 38.7\% higher PSNR, 18.1\% higher SSIM, and much lower FID, and is comparable to the large U-Net (24.5M), surpassing it on PSNR, SSIM, and LPIPS despite a slightly higher FID.
Our model's strong performance, especially in its lightweight version, comes from its hybrid architecture. By combining a Transformer Encoder for long-range spatial reasoning and FFC for efficient global context, the model captures both global structure and local details. The U-Net backbone integrates these features to produce accurate and coherent maps. This design enables \emph{SenseExpo} to achieve higher structural integrity and pixel-level accuracy than purely convolutional models with far fewer parameters.

\subsection{The Evaluation of Prediction Robustness} \label{sec:experiment:the_evaluation_of_prediction_robustness}
\begin{figure}[thpb]
	\centering
	\includegraphics[width=0.35\textwidth]{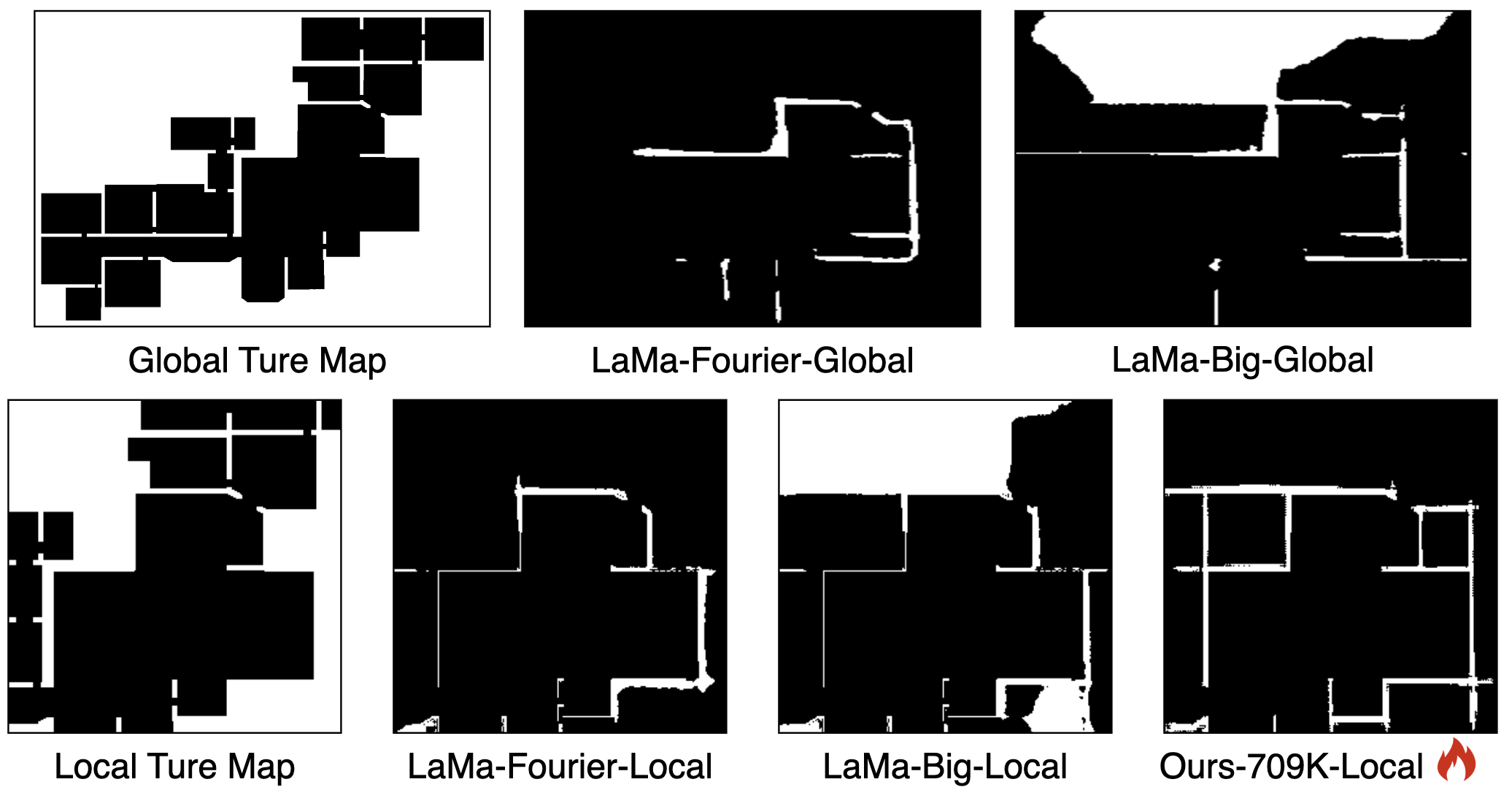}
	\caption{\textbf{Robustness of the models.} The models are tested on HouseExpo dataset~\cite{li2020houseexpo}. Our 709K model achieved SOTA by having better accuracy compared to LaMa models.}
	\label{fig8}
\end{figure}

To evaluate the robustness of prediction methods under domain shift, experiments were conducted on the HouseExpo dataset~\cite{li2020houseexpo}, a previously unseen environment, by constructing both local and global map prediction tasks. All models, exclusively trained on the KTH dataset~\cite{aydemir2012can}, were subjected to cross-domain testing to simulate real-world generalization challenges.
As shown in Fig.~\ref{fig8} and Tab.~\ref{tab:comprehensive_comparison}, the local prediction paradigm consistently outperformed its global counterpart, with Big LaMa-Fourier achieving a 32.6\% FID reduction (397.36 to 188.12) when moving from global to local inference. This suggests that localized feature extraction better mitigates domain discrepancies in unfamiliar environments.

Notably, our compact model (709K parameters) achieved superior robustness despite being far smaller than the LaMa variants (27M--51M). With an FID of 161.55 and LPIPS (Alex) of 0.377---14.1\% and 10.7\% below Big LaMa-Fourier's local predictions---it challenges the assumption that model capacity directly correlates with cross-domain robustness.

\subsection{Ablation Studies}
\label{subsec:ablation_main}

We assess SenseExpoNet on KTH using four variants: the U-Net baseline, U-Net+FFC, U-Net+Transformer, and the full model with both modules.
The Transformer branch contributes most of the PSNR and SSIM improvement, whereas removing both modules severely harms perceptual quality. Combining FFC with the Transformer restores low FID and surpasses the baseline without increasing parameters.
The number and placement of upsampling blocks are kept fixed across variants to isolate the two global-context modules. We do not claim this ablation exhaustively optimizes the discriminator loss or every decoder block placement; it is intended to show the contribution of Transformer and FFC components under the same lightweight backbone.

\begin{table}[t]
	\centering
	\footnotesize
	\setlength{\tabcolsep}{4pt}
	\renewcommand{\arraystretch}{0.95}
	\caption{Ablation study on the KTH dataset (mean over five seeds).}
	\label{tab:ablation_main}
	\begin{tabular*}{0.95\linewidth}{@{\extracolsep{\fill}}lcccc}
		\toprule
		Configuration & PSNR$\uparrow$ & SSIM$\uparrow$ & \makecell{LPIPS\\(Alex)$\downarrow$} & FID$\downarrow$ \\
		\midrule
		U-Net (Baseline, 1.1M) & 7.707 & 0.670 & 0.326 & 56.904 \\
		-- FFC & 7.945 & 0.683 & 0.274 & 81.039 \\
		-- Transformer \& FFC & 7.804 & 0.681 & 0.396 & 155.614 \\
		\textbf{Ours (FFC + Transformer)} & \textbf{9.026} & \textbf{0.718} & \textbf{0.283} & \textbf{42.353} \\
		\bottomrule
	\end{tabular*}
\end{table}

\begin{table}[h]
	\centering
	\setlength{\tabcolsep}{4pt}
	\caption{Deployment efficiency comparison on RTX 4090, RTX 5000, and Jetson AGX Orin (32~GB).}
	\label{tab:efficiency}
	\resizebox{\linewidth}{!}{%
		\begin{tabular}{@{}lccccc@{}}
			\toprule
			Device                           & Compute & Runs & Mean (ms)$\downarrow$    & Median (ms)$\downarrow$  & Min / Max (ms)$\downarrow$\\
			\midrule
			\textbf{Jetson AGX Orin (32 GB)} & CPU     & 10   & 332.01         & 329.69         & 289.14 / 383.85 \\
			\textbf{Jetson AGX Orin (32 GB)} & GPU     & 10   & \textbf{62.87} & \textbf{62.87} & 62.50 / 63.26   \\
			RTX 4090                         & GPU     & 10   & 6.21           & 6.19           & 6.05 / 6.34     \\
			RTX 5000                         & GPU     & 10   & 7.53           & 7.51           & 7.46 / 7.67     \\
			\bottomrule
		\end{tabular}%
	}
\end{table}
\begin{figure}[h]
	\centering
	\includegraphics[scale=0.43]{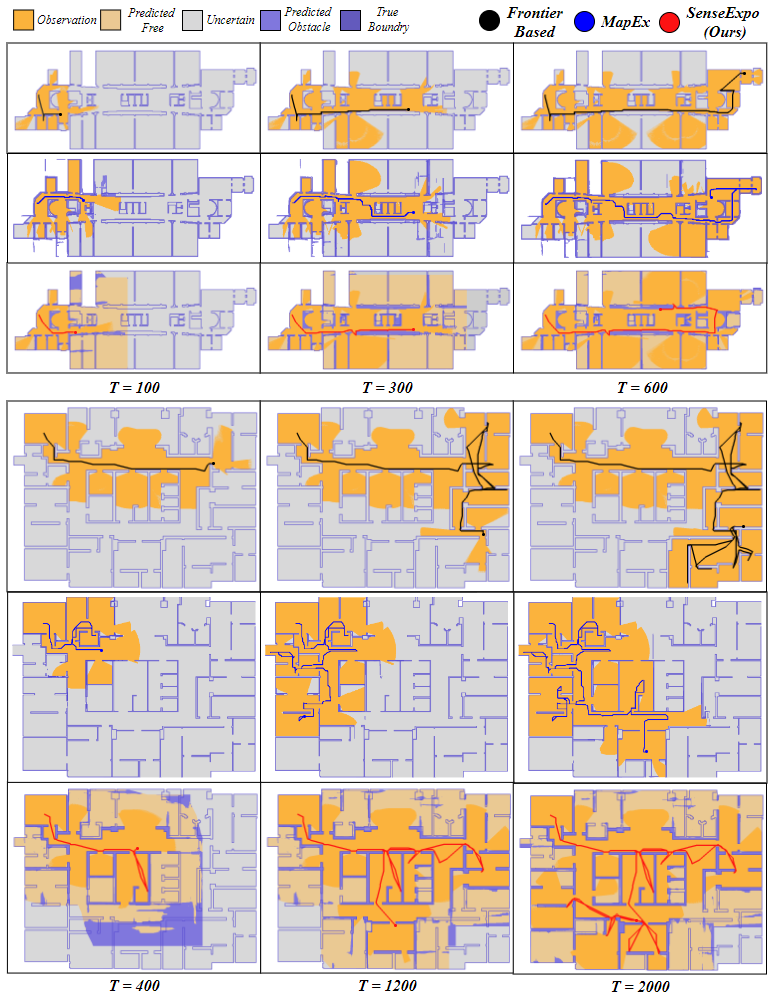}
	\caption{\textbf{Comparison of exploration efficiency in KTH and MRPB 1.0 Dataset.} When T reaches 500, our model has completed the prediction of the entire map, with no remaining uncertain areas, which are represented in gray.}
	\label{fig9}
	\vspace{-10pt}
\end{figure}

\begin{figure}[h]
	\centering
	\includegraphics[scale=0.75]{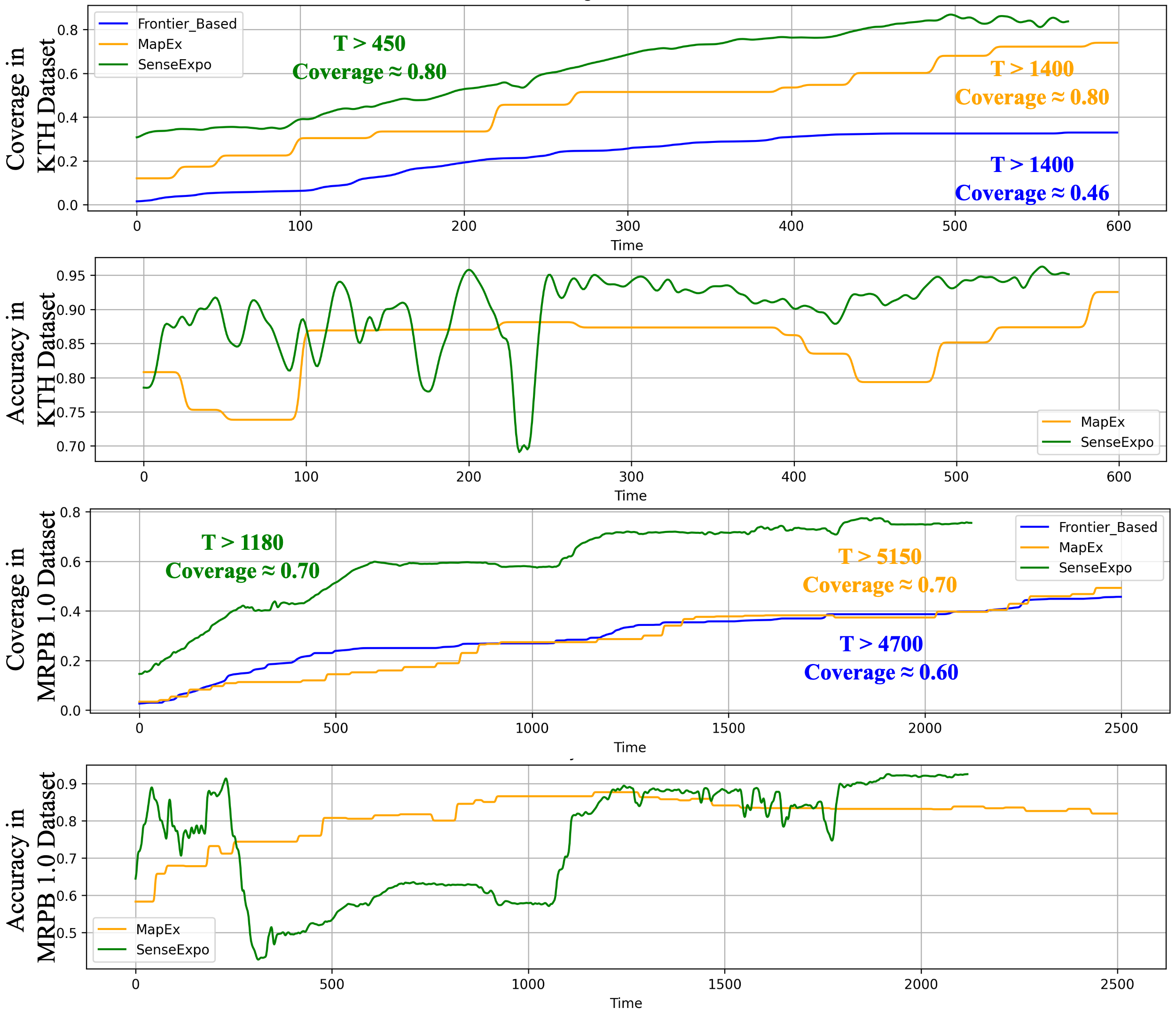}
	\caption{\textbf{Comparison of models in terms of coverage and accuracy.} Our model demonstrated higher coverage and accuracy in both KTH and MRPB 1.0 datasets. }
	\label{fig10}
	\vspace{-10pt}
\end{figure}

\subsection{The Comparison of Exploration Efficiency} \label{sec:experiment:the_comparation_of_exploration_efficiency}
We compared the exploration efficiency of the classic Frontier-based exploration method and the \emph{SenseExpo} exploration method with our 709k model in KTH dataset and MRPB 1.0. Additionally, we used the MapEx~\cite{ho2024mapex} method with 3 big LaMa models (51M) for exploration on the same scenarios and compared it with our exploration algorithm. It is important to note that all models in this experiment were trained solely on the KTH dataset, and \emph{SenseExpo} predicts free space while MapEx predicts obstacle structures. The results are presented in Fig.~\ref{fig9}, and Fig.~\ref{fig10}.
For the purpose of comparing efficiency, each exploration run was terminated when the respective algorithm could no longer find any new frontiers, signifying the end of the exploration process. The total time taken to reach this state was recorded.

As shown in Fig.~\ref{fig9}, in the KTH dataset, our model consistently exhibited higher coverage than both the frontier-based and MapEx models across all timestamps, with the gap widening over time. Our model reaches the same coverage as the classic frontier-based model with an approximately 92.5\% time reduction. Compared to MapEx, it reaches 80\% coverage with 67.9\% less time. Despite early fluctuations, our model maintained over 90\% accuracy throughout most of the evaluation, ultimately improving by 2.5\%. On the MRPB 1.0 dataset, our model achieves 60\% and 70\% coverage with around 77.1\% less time than the Frontier-based and MapEx models, respectively, highlighting its efficiency on large maps. Despite a temporary accuracy dip during exploration, it ultimately outperformed MapEx with an approximately 8\% accuracy gain.
The drastic reduction in exploration time stems from our model's high-quality predictions, which enable long-horizon goal selection rather than the incremental, myopic behavior of classic frontier-based methods. By predicting expansive regions of free space, SenseExpo lets the planner identify goals deep within unexplored territory; proactively identifying where the robot \emph{can} go is more efficient than predicting where it cannot (as in MapEx), yielding more direct trajectories and faster coverage.
For real-world deployment, we further evaluate the inference latency of SenseExpoNet on both desktop and embedded hardware, the results are provided in ~\ref{tab:efficiency}.

\section{CONCLUSION} \label{sec:conclusion}

In this work, we introduce \emph{SenseExpo}, a novel approach to enhance the efficiency and scalability of autonomous exploration. By leveraging lightweight neural networks, it balances high prediction accuracy with low computational overhead, making it suitable for resource-constrained robots. Our experiments show that \emph{SenseExpo} achieves SOTA performance across different datasets, and its ease of integration via ROS further contributes to practical applicability in real-world robotic exploration.
Future work includes improving robustness in highly dynamic environments and extending capabilities with RGB inputs and more diverse spatial layouts.


\section*{ACKNOWLEDGMENT}
This work was supported in part by the National Natural Science Foundation of China (U25A20489, 62334006), the Beijing Natural Science Foundation (L253009), the National Science and Technology Major Project Fund of China (2025ZD0215600), and the National Key Technologies R\&D Program of China (2025YFF1500600). The authors also thank D-Robotics for their support. The work of H. Que, T. Zhu, and H. Yao was carried out entirely in the United States and did not receive any funding.

\bibliographystyle{IEEEtran}
\bibliography{main}

\end{document}